\documentclass[letterpaper]{article} 
\usepackage{aaai2026}  
\usepackage{times}  
\usepackage{helvet}  
\usepackage{courier}  
\usepackage{multirow}
\usepackage{booktabs}
\usepackage{amsmath}
\usepackage{amsthm}
\usepackage{amssymb}
\usepackage{xcolor}
\usepackage[hyphens]{url}  
\usepackage{graphicx} 
\usepackage{natbib}  
\usepackage{caption} 
\usepackage{algorithm}
\usepackage{algorithmic}

\newtheorem{theorem}{Theorem}
\newtheorem{lemma}{Lemma}

\newcommand{\mc}{\mathcal}
\newcommand{\defeq}{\mathrel{\mathop:}=}

\newcounter{checksubsection}
\newcounter{checkitem}[checksubsection]

\usepackage{newfloat}
\usepackage{listings}
\DeclareCaptionStyle{ruled}{labelfont=normalfont,labelsep=colon,strut=off} 
\floatstyle{ruled}
\newfloat{listing}{tb}{lst}{}
\floatname{listing}{Listing}

\nocopyright

\title{Semi-Supervised Semantic Segmentation via Derivative Label Propagation}
\author{
    Yuanbin Fu, \textsuperscript{\rm 1}
    Xiaojie Guo \textsuperscript{\rm 1}\thanks{Corresponding author}\\
}
\affiliations{
    \textsuperscript{\rm 1}College of Intelligence and Computing,
    Tianjin University\\
}

\usepackage{bibentry}

\begin{document}

\maketitle

\begin{abstract}
Semi-supervised semantic segmentation, which leverages a limited set of labeled images, helps to relieve the heavy annotation burden. While pseudo-labeling strategies yield promising results, there is still room for enhancing the reliability of  pseudo-labels. Hence,  we develop a semi-supervised framework, namely DerProp, equipped with a novel derivative label propagation  to rectify imperfect pseudo-labels. Our label propagation method imposes discrete  derivative operations on pixel-wise feature vectors as additional regularization,  thereby generating strictly regularized similarity metrics. Doing so effectively alleviates the ill-posed problem that identical similarities correspond to different features, through constraining the solution space.  Extensive experiments are conducted to verify the rationality of our design, and demonstrate our superiority over other methods. Codes are available at \url{https://github.com/ForawardStar/DerProp/}.
\end{abstract}


\section{Introduction}
Semantic segmentation (SS) aims at segmenting the complete objects that belong to different semantic categories, which is instrumental  across various application scenarios, such as autonomous driving \cite{DBLP:conf/cvpr/QuJZYZ21,DBLP:journals/tits/FengHRHGTWD21,DBLP:conf/aaai/HanWHGSG25}, medical imaging \cite{DBLP:conf/ijcai/WangLL0K22,DBLP:conf/aaai/LiYWH20,DBLP:conf/miccai/XingYYLZ24,DBLP:conf/miccai/XingYWHZ22}, and robotic navigation \cite{DBLP:conf/miua/SandersonM22,DBLP:conf/iisa/TzelepiT21,DBLP:conf/iccv/LiZMQ023,DBLP:conf/cvpr/HanGPLGY25,DBLP:conf/prcv/DongWFDH23,DBLP:journals/tgrs/FanWHLDZLHD24}. With the rapid development of deep learning techniques, the accuracy of semantic segmentation has been evidently boosted, benefiting from their capacity to directly learn mappings from raw images to segmentation maps.  Unfortunately, manually annotating dense segmentation maps is time-consuming and laborious, presenting a significant barrier to the widespread collection of high-quality data. To relieve such annotation burden, semi-supervised semantic segmentation (SSSS)  has garnered substantial interest within the research community. Hence, the goal of SSSS is to attain the high segmentation accuracy,  with the benefit of solely employing a small amount of manually annotated samples alongside a large number of unlabeled ones.

\begin{figure}[t]
    \centering
    \includegraphics[width=0.298\linewidth]{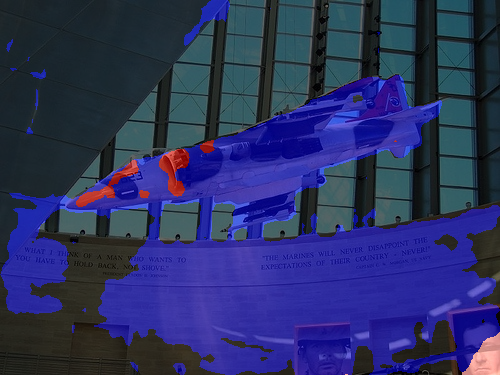}
    \includegraphics[width=0.298\linewidth]{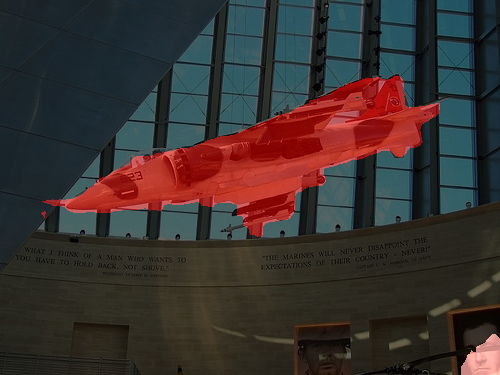}
    \includegraphics[width=0.298\linewidth]{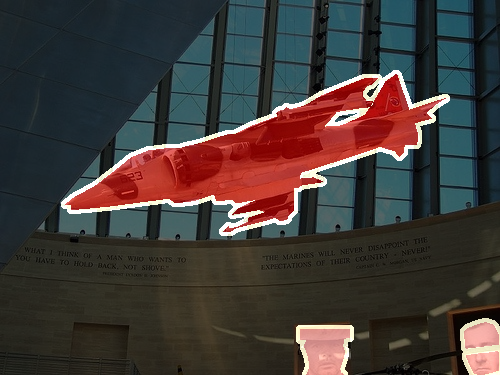}

    \includegraphics[width=0.298\linewidth]{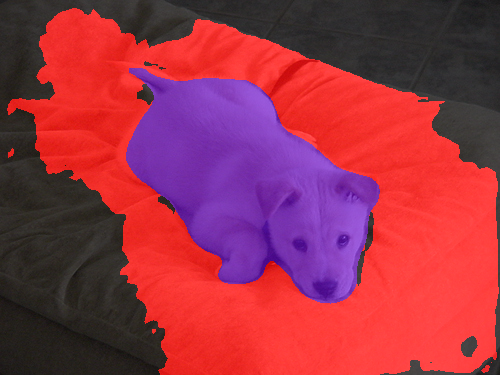}
    \includegraphics[width=0.298\linewidth]{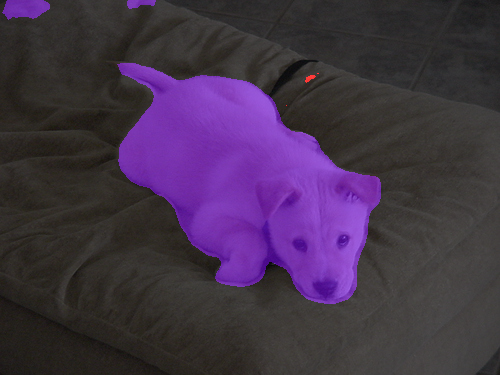}
    \includegraphics[width=0.298\linewidth]{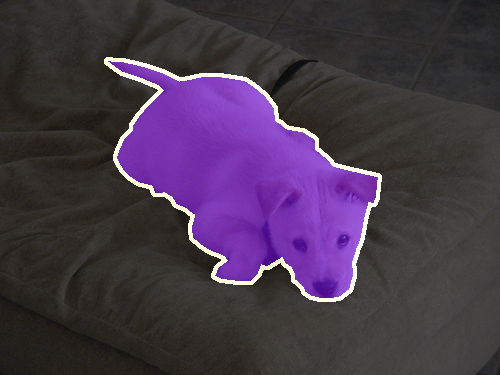}

    \includegraphics[width=0.298\linewidth]{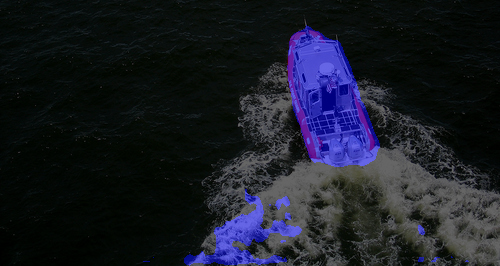}
    \includegraphics[width=0.298\linewidth]{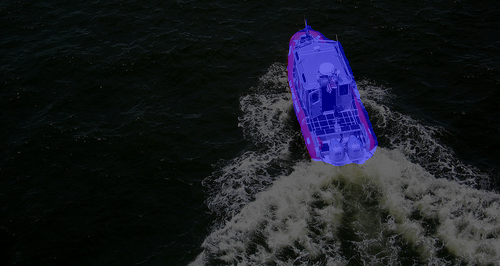}
    \includegraphics[width=0.298\linewidth]{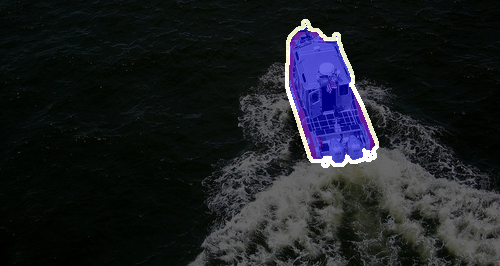}

     w/o DLP \;\quad\quad\quad w/ DLP \;\quad\quad\quad\quad\; GT \quad
    
    \caption{Visual comparisons between with and without our proposed derivative label propagation (DLP).}
\label{first_pic}
\end{figure}

Over the past decades, a plethora of SSSS methodologies have emerged, which predominantly follow a pipeline of generating pseudo-labels by deep neural networks for those unlabeled samples. In particular, existing SSSS methods can be roughly divided into: 1) mean teacher based methods \cite{MeanTeacher1,MeanTeacher2,MeanTeacher3,MeanTeacher4} that accumulate the network parameters updated at different training iterations to produce pseudo-labels, 2) consistency regularization based methods \cite{DBLP:journals/tip/KeAPRS22,DBLP:conf/cvpr/YangZ0S022,DBLP:conf/iccv/YuanLSW021,CorrMatch} that enforce the predictions corresponding to weakly and strongly augmented \cite{DBLP:conf/mm/LinZXWWDDC24,DBLP:journals/spl/LinWWLG24} inputs to be consistent, and 3) confidence thresholding based methods \cite{MeanTeacher3,CorrMatch} that set thresholds to preserve reliable pseudo-labels and suppress unreliable pseudo-labels. However,  existing methods still suffer from  producing imperfect pseudo-labels, hindering further accuracy improvements. To be more specific, the pseudo-labeling process inevitably suffers from noises/outliers, \textit{i.e.}, incorrect labels, which severely misguide the training optimization direction. Consequently, networks trained on noisy pseudo-labels may generate noisier labels in subsequent iterations, leading to the error accumulation problem. 

For the sake of improving the performance of SSSS, the pseudo-labels ought to be further rectified so that their reliability can be guaranteed. But, it is challenging because there are no accurate dense labels available for supervising the rectification process. In other words, the ground truth conversions from wrong pseudo-labels to desired true labels is unknown. To address this issue, it is crucial to excavate valuable information  inherent within the data itself or the learning process. Therefore, based on the principle that pixels with high similarities are highly  likely to share the same class label while pixels with low similarities may differ in semantic category, the label propagation technique \cite{CorrMatch,DBLP:conf/cvpr/StojnicKMT25,DBLP:conf/iccv/PapadopoulosW021,DBLP:conf/eccv/ZhangDJZ20} can be leveraged to rectify the misclassified pixels based on semantic similarities with neighboring pixels. This strategy, however, raises a critical question: \emph{how can we reliably measure the similarities among pixels, so as to accurately identify and rectify the potential misclassified pixels in pseudo-labels?}

\noindent\textbf{Contributions}. In this paper, we answer the above question by developing a  semi-supervised framework, named DerProp. The proposed framework adopts a simple yet effective derivative label propagation (DLP) method, which imposes the discrete  derivative operation on the pixel-wise feature vectors to obtain derivative-based features.   Specifically, besides supervising/regularizing the similarities between the original feature vectors, the similarities between the derivative feature vectors should also be properly regularized. By doing so, the ill-posed problem that the same similarity scores may correspond to multiple (wrong) solutions, can be alleviated. For example, consider a simple case: $[1, 1, 1]$ and $[2 + \sqrt{3}, 1, 0]$ are the 3-dimensional feature vectors that can correctly represent the semantics of two different pixels. The cosine similarity between $[1, 1, 1]$ and $[2 + \sqrt{3}, 1, 0]$, is the same as that between another  pair of vectors, \textit{i.e.}, $[1, 1, 1]$ and $[2 - \sqrt{3}, 1, 0]$, both being $\sqrt{2}/2$. But, only $[2 + \sqrt{3}, 1, 0]$  is the correct one. Using $\sqrt{2}/2$ as ground truth to supervise the similarity between these two pixels, encounters a high risk of generating the incorrect $[2 - \sqrt{3}, 1, 0]$ to represent the pixel, leading to misclassification for it. To resolve such ill-posedness, derivative-based features can be employed as additional constraints for restricting the solution space. 

Therefore, we propose to calculate the similarities with respect to original feature vectors together with additional derivative feature vectors. Theoretical analysis is provided to formally reveal how discrete  derivative operators alleviate the ill-posedness. Further, to mitigate the computational expense of high-order ( $\geq$ 2) discrete  derivative operations, we introduce a sparsity term to approximate the regularization on the high-order derivative-based features. As shown in Fig. \ref{first_pic}, our DLP can evidently improve the segmentation performance through introducing additional regularization.

Our contributions can be summarized as:
\begin{itemize}
    \item  We propose a novel derivative label propagation method that performs discrete  derivative operations on the feature vectors. It can alleviate the ill-posed problem that identical similarities may result in different features.
    \item We prove that high-order discrete  derivative operations on feature vectors make the problem well-posed, while the sparsity regularization on the 2-order derivative-based features can provide a well approximation.
    \item We conduct extensive experiments to quantitatively and qualitatively verify the effectiveness of our design on several benchmark datasets, and demonstrate our superiority over other state-of-the-art methods.
\end{itemize}

\section{Related Work}
\noindent\emph{Fully-Supervised Semantic Segmentation}.
 The evolution of fully-supervised segmentation can be traced back to Fully Convolutional Networks (FCNs) \cite{FCN}, which pioneers the integration of end-to-end deep neural networks.  Subsequent advancements in enlarging the receptive fields are addressed through Atrous Spatial Pyramid Pooling (ASPP) \cite{ASPP}. HRNet \cite{HRNet} enhances the high-resolution representations through aggregating the up-sampled representations from parallel convolutions. Zhang \textit{et al.} \cite{DBLP:conf/cvpr/0005DSZWTA18} explored the influence of global contextual information through designing a novel context encoding module. SegNet \cite{SegNet} is composed of a series of encoding layers followed by a decoding layer, which maps the low-resolution features to high-resolution features. To accelerate the running speed, BiSeNet \cite{BiSeNet} uses a spatial branch with small-stride convolutions to preserve fine-grained details through high-resolution feature extraction. Despite achieving promising performance, these methods require manually annotated dense segmentation labels for all training samples, which are expensive to collect.

\noindent\emph{Semi-Supervised Semantic Segmentation}. 
Generally, the mainstream solution for semi-supervised segmentation is to generate pseudo-labels using deep neural networks for those unlabeled samples, which can be roughly divided into: 1) mean teacher, 2) consistency regularization, and 3) confidence thresholding based methods.

Specifically, for the mean teacher based methods, AEL \cite{MeanTeacher1} incorporates a confidence bank that dynamically tracks per-category performance, enabling adaptive re-weighting of training focus between strong and weak categories. GTA-Seg \cite{MeanTeacher2} explicitly disentangles the effects of pseudo-labels on the feature extraction and segmentation prediction pathways. Several recent works \cite{MeanTeacher3,MeanTeacher4} also introduced novel extensions to the mean teacher model for enhancing prediction accuracy. Another technological route is the consistency learning based methods that enforce the predictions corresponding to diversely perturbed inputs to be consistent. As representatives, the key idea of the work by \cite{DBLP:journals/tip/KeAPRS22} involves extracting pseudo masks on unlabeled data coupled with multi-task segmentation consistency enforcement. ST++ \cite{DBLP:conf/cvpr/YangZ0S022} applies strong data augmentations to unlabeled images, which mitigates noisy label overfitting and decouples teacher-student predictions. The distribution-specific batch normalization designed by \cite{DBLP:conf/iccv/YuanLSW021} helps to address the problem of large distribution disparities caused by strong augmentation. Besides the aforementioned methods, many researchers have explored the influences of confidence thresholding in SSSS.  For instance, CorrMatch \cite{CorrMatch} generates a binary map by thresholding confidence scores from weakly augmented inputs, explicitly identifying trustworthy pixels in pseudo-labels for semi-supervised training.

Despite achieving promising results, the quality of pseudo-labels continues to limit the accuracy. We overcome this limitation through a novel derivative label propagation method that rectifies unreliable pseudo-labels.

\section{Methodology}
The labeled training set is: $\mathcal{D}_l\defeq \{(\mathbf{X}_{nl}\in\mathbb{R}^{3\times M_{nl}}, \mathbf{Y}_{nl}\in\mathbb{R}^{C\times M_{nl}}, nl \defeq 1, 2, ..., N_l \}$, and the unlabeled set is: $\mathcal{D}_{u} \defeq \{\mathbf{X}_{nu}\in\mathbb{R}^{3\times M_{nu}}, nu \defeq 1, 2, ..., N_u \}$, where $N_l$ and $N_u$ are the number of labeled and unlabeled samples, respectively. The total pixel amount of the $nl$-th labeled and $nu$-th unlabeled image, and the number of classes, are $M_{nl}$,  $M_{nu}$ and $C$, respectively. Note that images are represented as flattened 2D matrices in this paper, \textit{i.e.}, $\mathbb{R}^{\text{channel} \times \text{pixel}}$, for notational simplicity, originating from their original 3D structure, \textit{i.e.}, $\mathbb{R}^{\text{channel} \times \text{height} \times \text{width}}$, where pixel = height $\times$ width. During training, weak and strong augmentations are performed on $\mathbf{X}_{nu}$, to obtain $\mathbf{X}^w_{nu}$ and  $\mathbf{X}^s_{nu}$, respectively. The predictions from $\mathbf{X}^w_{nu}$, \textit{e.g.}, $\mathbf{P}^w_{nu} \defeq \text{Softmax}(\mathcal{G}(\mathbf{X}^w_{nu}))$, can supervise the predictions from $\mathbf{X}^s_{nu}$, \textit{e.g.}, $\mathbf{P}^s_{nu} \defeq \text{Softmax}(\mathcal{G}(\mathbf{X}^s_{nu}))$, where $\mathcal{G}(\cdot)$ is the segmentation network.

\subsection{Derivative Label Propagation}
Generally,  the label propagation can be described as:
\begin{equation}
  \widetilde{\mathbf{L}}^w_{nu}  \defeq  \mathbf{L}^w_{nu}  \mathbf{S}_{nu},
\end{equation}
where $\mathbf{L}^w_{nu} \defeq \mathcal{G}(\mathbf{X}^w_{nu}) \in\mathbb{R}^{C\times M_{nu}}$ represents the class logits, and $\widetilde{\mathbf{L}}^w_{nu}\in\mathbb{R}^{C\times M_{nu}}$ is the rectified logits after performing the label propagation. The rectified pseudo-labels can thus be obtained by: $\widetilde{\mathbf{P}}^w_{nu} \defeq \text{Softmax}(\widetilde{\mathbf{L}}^w_{nu})$. $\mathbf{S}_{nu} \in \mathbb{R}^{M_{nu} \times M_{nu}}$ is the similarity matrix that models the similarities among pixels. The $i$-th row and $j$-th column in $\mathbf{S}_{nu}$ is calculated by $\mathcal{S}(\mathbf{v}_{nu,i}, \mathbf{v}_{nu,j})$, where  $\mathcal{S}(\cdot, \cdot)$ means the cosine similarity, and $\mathbf{v}_{nu,i}\in\mathbb{R}^D$ is the feature vector representing the semantics of the $i$-th pixel. For concise expression, we omit the subscript $nu$ in some following paragraphs, \textit{e.g.}, $M_{nu} \rightarrow M$,  $\mathbf{S}_{nu} \rightarrow \mathbf{S}$, and $\mathbf{v}_{nu,i} \rightarrow \mathbf{v}_i$. 

However,  solely supervising the similarity between the original $D$-dimensional feature vectors during training, may suffer from the ill-posed problem that identical similarities correspond to different solutions  (including those incorrect/biased solutions).
Hence, to make the problem well-posed so that extra incorrect solutions can be avoided, we propose to further impose the discrete  derivative operation:
\begin{equation}
\Delta^q \mathbf{v}(i) \defeq  \Delta^{q-1} \mathbf{v}(i+1) - \Delta^{q-1} \mathbf{v}(i),
\label{eq:derivative1}
\end{equation}
where $\Delta^q  \mathbf{v}\in\mathbb{R}^{D_q}$ represents the $D_q$-dimensional derivative feature vectors calculated through the $q$-order discrete  derivative operation ($q \in \{1, 2, ..., D - 1 \}$), and $D_q = D_{q-1} -1$. $\Delta^q \mathbf{v}(i)$ means the $i$-th element of the vector $\Delta^q \mathbf{v}$.  The values in original features $\mathbf{v}$ will be normalized so that $||\mathbf{v}||_1 = 1$, where $||\cdot||_1$ is L1 norm. When the order of the discrete  derivative operation is zero, we have: $\Delta^0 \mathbf{v} = \mathbf{v}$. We emphasize that rather than the spatial dimension of features, our proposed derivative operations are actually performed along the channel dimension to calculate the divergences among different features.
 The ill-posed problem can therefore be addressed based on the following theorem.

\begin{figure}[t]
    \centering
    
    \includegraphics[width=\linewidth]{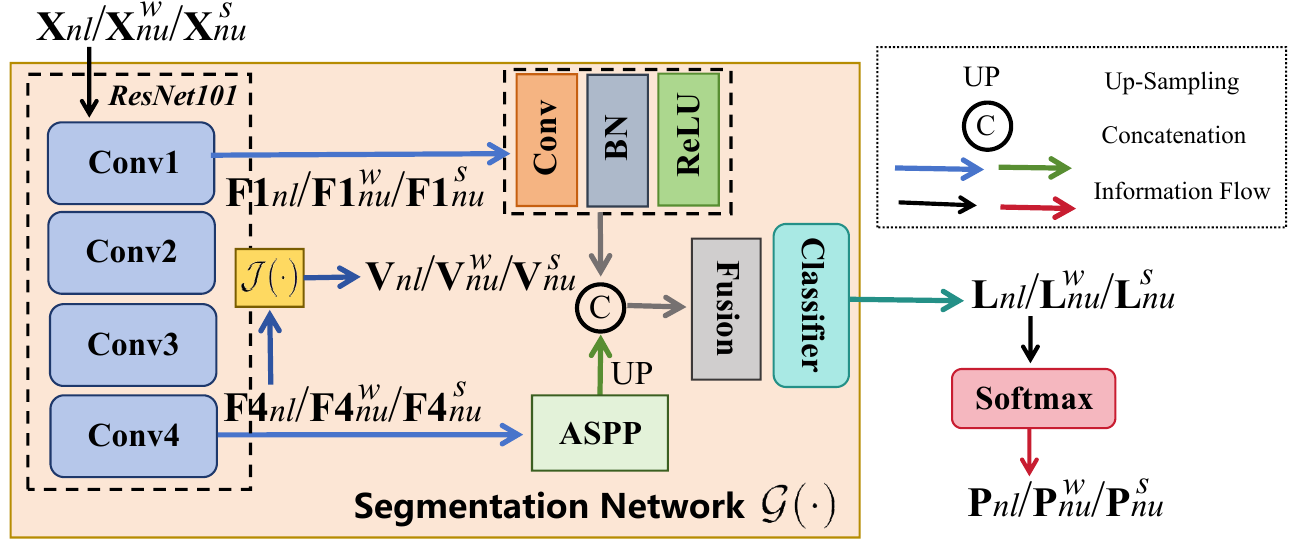}
    
    \caption{Network architecture of $\mathcal{G}(\cdot)$. The symbol `/' means different inputs or predictions. For example,  `$\mathbf{X}_{nl}/\mathbf{X}^w_{nu}/\mathbf{X}^s_{nu}$' means that the input to the network may be one of $\mathbf{X}_{nl}$, $\mathbf{X}^w_{nu}$, or $\mathbf{X}^s_{nu}$, while `$\mathbf{P}_{nl}/\mathbf{P}^w_{nu}/\mathbf{P}^s_{nu}$' denotes the prediction by $\mathcal{G}(\cdot)$  corresponding to $\mathbf{X}_{nl}$, $\mathbf{X}^w_{nu}$, or $\mathbf{X}^s_{nu}$, respectively.}
\label{arch}
\end{figure}

\begin{theorem}[Well-Posedness]
Assuming the feature vectors are L1-normalized, \textit{i.e.}, $||\mathbf{v}||_1 = 1$, there exists a unique solution for the pixel-wise feature vectors $\mathbf{v}_i$ ($i \defeq 1, 2, ..., M$), only if: 1) $\mathcal{S}(\mathbf{v}_{i}, \mathbf{v}_{j}) = s_{i,j}$, and 2) $\mathcal{S}(\Delta^q \mathbf{v}_{i}, \Delta^q \mathbf{v}_{j}) =  s^q_{i,j}, \forall q \in\{1, 2, ..., D-1\}$. Among them, $s_{i, j}$ and $s^q_{i,j}$ are the supervisions on the similarities with respect to  original feature vectors and derivative feature vectors, respectively.
\label{theorem1}
\end{theorem}
 Unfortunately,  $\forall q\in\{1, 2, ..., D-1 \}$, sequentially performing the $q$-order discrete  derivative operation is computationally expensive. Hence,  we employ an approximation of the regularization on high-order derivative-based features:
\begin{equation}
\begin{aligned}
    \mc{L}^{Der} \defeq & ||\mathbf{S} - \mathbf{S}^{GT}||_1 + ||\Delta^1 \mathbf{S} - \Delta^1 \mathbf{S}^{GT}||_1 \\
    + & \eta ||\Delta^2 \mathbf{V}||_1,
    \end{aligned}
    \label{eq:contrastive_diff}
\end{equation}
where $\eta$ is a balancing weight. Given $||\mathbf{v}_i||_1=1$, the cosine similarity satisfies:  $\mathcal{S}(\mathbf{v}_{i}, \mathbf{v}_{j})\defeq \mathbf{v}_i^{\top}\mathbf{v}_j/||\mathbf{v}_i||_1||\mathbf{v}_j||_1= \mathbf{v}_i^{\top}\mathbf{v}_j$. Thus, the $i$-th row and $j$-column in $\mathbf{S}\defeq \mathbf{V}^{\top}\mathbf{V}\in\mathbb{R}^{M\times M}$ and $\Delta^1 \mathbf{S}\defeq (\Delta^1\mathbf{V})^{\top}(\Delta^1\mathbf{V})\in\mathbb{R}^{M\times M}$, are $\mathcal{S}(\mathbf{v}_{i}, \mathbf{v}_{j})$ and $\mathcal{S}(\Delta^1 \mathbf{v}_{i}, \Delta^1 \mathbf{v}_{j})$, respectively. The feature $\mathbf{V}$ is calculated by: $\mathbf{V} \defeq \mathcal{J}(\mathbf{F}4)$, where $\mathcal{J}(\cdot)$ is the projection layer (the orange rectangle in Fig. \ref{arch}), and $\mathbf{F}4$ means the features generated by the 4-th block of our backbone, \textit{i.e.}, ResNet101. The $i$-th column in $\mathbf{V}\in\mathbb{R}^{D\times M}$, $\Delta^1 \mathbf{V}\in\mathbb{R}^{(D-1)\times M}$  and  $\Delta^2 \mathbf{V}\in\mathbb{R}^{(D-2)\times M}$, are $\mathbf{v}_{i} \in\mathbb{R}^{D}$, $\Delta^1 \mathbf{v}_i\in\mathbb{R}^{D-1}$, and $\Delta^2 \mathbf{v}_{i}\in\mathbb{R}^{D-2}$, respectively.  $\mathbf{S}^{GT}$ and  $\Delta^1 \mathbf{S}^{GT}$ are the ground truth similarity matrices, the $i$-th row and $j$-th column of which are $s_{i,j}$ and $s^1_{i,j}$, respectively.

We can infer from \textbf{Theorem} 2 that,  minimizing $||\Delta^2 \mathbf{V}||_1$, is equivalent to minimizing  $||\Delta^q \mathbf{S} - \mathbf{0}||_1$ since $(2^{q-2} ||\Delta^2\mathbf{V}||_1)^2$ is always greater than $||\Delta^{q}\mathbf{S} - \mathbf{0}||_{1} $, where $\textbf{0}$ is an all-zero matrix.

\begin{theorem}[Boundness]
Assuming feature vectors are bounded, \textit{i.e.}, $||\mathbf{v}||_1 \leq 1$, $\forall q \in\{2, ..., D-1\}$, the similarity matrix $\Delta^{q}\mathbf{S}$ satisfies:
$
||\Delta^{q}\mathbf{S} - \mathbf{0}||_{1} \leq (2^{q-2} ||\Delta^2\mathbf{V}||_1)^2
$.
\label{theorem2}
\end{theorem}


    

 Since the similarities with respect to high-order ($q > 1$) derivative-based features are enforced to be 0 for all pixels,  their functionality in indicating the pixel-level semantic similarities, is suppressed. The key role of our sparsity regularization on these high-order derivative-based features, is to address the ill-posedness according to \textbf{Theorem} \ref{theorem1}.   Hence, we perform the  label propagation for rectifying the pseudo-labels of those unlabeled samples through:
\begin{equation}
  \widetilde{\mathbf{L}}^w_{nu}  \defeq  \mathbf{L}^w_{nu}  (\mathbf{S}^w_{nu} + \Delta^1 \mathbf{S}^w_{nu}),
  \label{DLP}
\end{equation}
where  $\mathbf{S}^w_{nu} \defeq \mathbf{V}_{nu}^{w\top}\mathbf{V}^w_{nu}$, and $\mathbf{V}^w_{nu}$ is projected by: $\mathbf{V}^w_{nu}\defeq \mathcal{J}(\mathbf{F}4^w_{nu})$ corresponding to the input $\mathbf{X}^w_{nu}$. The pseudo-labels can thus be obtained by:
\begin{equation}
    \overline{\mathbf{P}}_{nu} \defeq (1 - \eta_{ep}) \mathbf{P}^{w}_{nu} + \eta_{ep}  \widetilde{\mathbf{P}}^{w}_{nu},
    \label{eq:pseudolabel_refine}
\end{equation}
where $\mathbf{P}^{w}_{nu}\defeq \text{Softmax}(\mathbf{L}^{w}_{nu})$, and $\widetilde{\mathbf{P}}^{w}_{nu}\defeq \text{Softmax}(\widetilde{\mathbf{L}}^{w}_{nu})$. 
$\eta_{ep}$ ramps up as $ep/EP$ to address the poor performance of $\widetilde{\mathbf{P}}^{w}_{nu}$ at the early training period.

\subsection{Network Architecture}
As illustrated in  Fig. \ref{arch}, our segmentation network adopts the pre-trained ResNet101 as backbone. Following previous works \cite{FixMatch,UniMatch,CorrMatch,DBLP:conf/iclr/Zhang0BW025}, for  $\mathcal{G}(\cdot)$, the features from the 1-st and 4-th block of ResNet101 are fed into a Conv-BN-ReLU layer and ASPP module \cite{ASPP}, respectively. Then, these two groups of features are fused, and processed by a classifier to output the class logits, \textit{e.g.}, $\mathbf{L}_{nl}$, $\mathbf{L}^w_{nu}$, or                  $\mathbf{L}^s_{nu}$ are output by $\mathcal{G}(\mathbf{X}_{nl})$, $\mathcal{G}(\mathbf{X}^w_{nu})$, and $\mathcal{G}(\mathbf{X}^s_{nu})$, corresponding to inputs  $\mathbf{X}_{nl}$,     $\mathbf{X}^w_{nu}$, or $\mathbf{X}^s_{nu}$, respectively. 

Drawing inspiration from ensemble learning principles  \cite{ensembles,Generalization,RandomForests,DBLP:conf/emnlp/BaiZ0KG0K024}, biases/interferences of a single training moment can be compensated  by aggregating complementary information from multiple diverse sources. Hence, we aggregate knowledge across different training moments to yield better results than those derived from a single training moment.
Specifically,  we maintain a momentum network $\mathcal{G}^m(\cdot)$ corresponding to $\mathcal{G}(\cdot)$ that is trained by back-propagation. The parameters of our momentum network are updated by accumulating the parameters of its corresponding back-propagation network:
\begin{equation}
    \mathbf{\Theta}_{(ep)}^m \defeq (\mathbf{\Theta}_{(ep-1)}^m  + \mathbf{\Theta}_{(ep)}^b ) / 2,
    \label{momentum}
\end{equation}
where $\mathbf{\Theta}_{(ep)}^m$ and $\mathbf{\Theta}_{(ep)}^b$ are the parameters of the momentum network $\mathcal{G}^m(\cdot)$ and back-propagation network $\mathcal{G}(\cdot)$, after the $ep$-th epoch, respectively. During training, the pseudo-labels are generated by the back-propagation network. In testing, only the momentum network $\mathcal{G}^m(\cdot)$ is executed, while the back-propagation network  $\mathcal{G}(\cdot)$ is abandoned.

\subsection{Loss Function}
Our overall loss $\mathcal{L}$ consists of: cross entropy loss $\mathcal{L}^{CE}$, KL divergence loss $\mathcal{L}^{KL}$, and derivative loss $\mathcal{L}^{Der}$:
\begin{equation}
    \mathcal{L} \defeq  \lambda^{CE}  \mathcal{L}^{CE} + \lambda^{KL}  \mathcal{L}^{KL} + \lambda^{Der}  \mathcal{L}^{Der},
    \label{loss:segmentation}
\end{equation}
where $\lambda^{CE}$, $\lambda^{KL}$, and $\lambda^{Der}$ are balancing weights.

Specifically, denote by $l^{CE}$ the pixel-wise cross-entropy loss function, we have:  
$
    \mathcal{L}^{CE}\defeq  \sum_{nl\defeq 1}^{N_l} l^{CE}(\mathbf{P}_{nl}, \mathbf{Y}_{nl}) + \sum_{nl\defeq 1}^{N_l} l^{CE}(\widetilde{\mathbf{P}}_{nl}, \mathbf{Y}_{nl}) +
    [\sum_{nu\defeq 1}^{N_u} l^{CE}(\mathbf{P}^s_{nu}, \overline{\mathbf{P}}_{nu}) + \sum_{nu\defeq 1}^{N_u} l^{CE}(\widetilde{\mathbf{P}}^s_{nu}, \overline{\mathbf{P}}_{nu})]/2$. Among them, $\mathbf{P}_{nl}\defeq \text{Softmax}(\mathbf{L}_{nl})$, and $\mathbf{P}^s_{nu}\defeq \text{Softmax}(\mathbf{L}^s_{nu})$, with $\mathbf{L}_{nl}\defeq \mathcal{G}(\mathbf{X}_{nl})$, and $\mathbf{L}^s_{nu}\defeq \mathcal{G}(\mathbf{X}^s_{nu})$. $\widetilde{\mathbf{P}}_{nl}$ and $\widetilde{\mathbf{P}}^s_{nu}$ are obtained by: $\text{Softmax}(\mathbf{L}_{nl}  (\mathbf{S}_{nl} + \Delta^1 \mathbf{S}_{nl}))$ and $\text{Softmax}(\mathbf{L}^{s}_{nu}  (\mathbf{S}^s_{nu} + \Delta^1 \mathbf{S}^s_{nu}))$, respectively. Denote by $\mathbf{V}_{nl}$ and $\mathbf{V}^s_{nu}$ the features projected by $\mathcal{J}(\cdot)$ respectively corresponding to inputs $\mathbf{X}_{nl}$ and $\mathbf{X}^s_{nu}$, we have: $\mathbf{S}_{nl}\defeq \mathbf{V}_{nl}^{\top}\mathbf{V}_{nl}$ and  $\mathbf{S}^s_{nu}\defeq \mathbf{V}_{nu}^{s\top}\mathbf{V}^s_{nu}$. Different from $\mathcal{L}^{CE}$, the KL divergence loss is used only for unlabeled samples: $\mathcal{L}^{KL}\defeq  \sum_{nu\defeq 1}^{N_u} \text{KL}(\mathbf{P}^s_{nu}, \overline{\mathbf{P}}_{nu})$, where $\text{KL}(\cdot, \cdot)$ is Kullback-Leibler divergence function. Following previous SSSS methods \cite{UniMatch, CorrMatch}, a binary map selecting high-confident pixels in pseudo-labels, is introduced for both  $\mathcal{L}^{CE}$ and $\mathcal{L}^{KL}$.

For the derivative loss, we construct the ground truth similarity matrix through: $\mathbf{S}^{GT}_{nl}\defeq \mathbf{Y}^{\top}_{nl}\mathbf{Y}_{nl}$, $\mathbf{S}^{GT}_{nu}\defeq \mathbf{V}^{w\top}_{nu}\mathbf{V}^w_{nu}$, and $\mathbf{S}^{GT}_{nu}\defeq (\Delta^1\mathbf{V}^{w}_{nu})^{\top}(\Delta^1\mathbf{V}^w_{nu})$, where $\mathbf{Y}_{nl}\in\mathbb{R}^{C\times M_{nl}}$ is the manually annotated one-hot segmentation label, and $\mathbf{V}^w_{nu}\defeq\mathcal{J}(\mathbf{F}4^w_{nu})$ is the projected features corresponding to $\mathbf{X}^w_{nu}$. Hence,  the derivative loss is formulated as:
\begin{equation}
\begin{aligned}
    \mc{L}^{Der} &\defeq  \sum^{N_l}_{nl\defeq 1} (||\mathbf{S}_{nl} - \mathbf{S}_{nl}^{GT}||_1 + ||\Delta^1 \mathbf{S}_{nl} - \ \mathbf{S}_{nl}^{GT}||_1 \\
    + & \eta ||\Delta^2 \mathbf{V}_{nl}||_1) 
    +  \sum^{N_u}_{nu\defeq 1} (||\mathbf{S}^s_{nu} - \mathbf{S}_{nu}^{GT}||_1 +\\
    & ||\Delta^1 \mathbf{S}^s_{nu} - \Delta^1 \mathbf{S}_{nu}^{GT}||_1 
    +  \eta ||\Delta^2 \mathbf{V}^s_{nu}||_1).
    \end{aligned}
    \label{loss:derivative}
\end{equation}

\section{Experimental Validation}
\subsection{Implementation Details}
We implement our network using PyTorch library on two V100 GPUs. In detail,  the segmentation network $\mathcal{G}(\cdot)$ is randomly initialized and trained for 80 epochs using SGD optimizer, the input size of which is $321\times 321$. $\lambda^{CE}$, $\lambda^{KL}$, and $\lambda^{Der}$ are set to be 0.5, 0.25, and 0.5, respectively. $\eta$ is set to be 0.5. The learning rate,  momentum and weight decay of the SGD optimizer are set to 0.001, 0.9 and 0.0001, respectively. During training, for labeled images, we randomly adjust the size of the original image to half or twice its previous size, randomly crop a region of $321\times 321$ from the original image, and randomly flip the image horizontally, for data augmentation. For unlabeled images, besides the aforementioned augmentation operations, we further perform random color perturbations, randomly convert RGB images to grayscale images, randomly blur the image, and randomly mix two different images. The batch size is set to be 4, and the total number of training epochs is 80.

\begin{table*}[pt]
\centering
\begin{tabular}{c|c|c|c|ccccc}
\toprule
Method & Year & Input Size & Params. & 1/16 (92) & 1/8 (183) & 1/4 (366) & 1/2 (732)  & Full (1464)\\
\midrule
UniMatch & CVPR'23 &  $321^2$ &  59.5M &  75.2 & 77.2 & 78.8 & 79.9  & 81.2 \\
DDFP & CVPR'24 &  $513^2$ & 59.5M & 74.9 & 78.0 & 79.5 & 81.2 & 81.9 \\
AllSpark & CVPR'24 &  $513^2$ & 89.3M & 76.0 & 78.4 & 79.7 & 80.7 & 82.1 \\
RankMatch & CVPR'24 &  $513^2$ &  - &  75.5 & 77.6 & 79.8 & 80.7 & \textbf{82.2}  \\
CorrMatch & CVPR'24 &  $321^2$ &  59.5M &  76.4 & 78.5 & 79.4 & 80.6 & 81.8 \\
MGCT & TMM'25 &  $321^2$ &  - &  75.2 & 76.7 & 76.9 & -  & - \\
RCC & AAAI'25 &  $513^2$ &  - & 75.3  & 77.9 & 79.8 & 81.0 & 82.1 \\
ScaleMatch & AAAI'25 &  $321^2$ &  - &  76.1 & 78.6 & 79.6 & 80.7 & 81.8 \\
\midrule
DerProp w/o DLP & - & $321^2$  & 59.5M  &  66.6 & 77.8 & 78.3 & 79.8 & - \\
DerProp w/o $\mathcal{L}^{Der}$ & - & $321^2$  & 59.5M  &  72.9 & 77.0 & 77.5 & 78.8 & 81.0\\
DerProp w/o momentum & - & $321^2$  & 59.5M  & 75.1 & 78.0 & 79.6 & 78.8 & 81.9 \\
\midrule
DerProp (Ours) & - & $321^2$  & 59.5M  &  \textbf{77.6} & \textbf{78.7} & \textbf{80.5} & \textbf{81.3} & \textbf{82.2} \\
\bottomrule
\end{tabular}
\caption{Results on Pascal VOC 2021 in terms of mIoU (\%), with ResNet101 as the backbone. Best results are \textbf{Bolded}.}
\label{PASCALVOC_result}
\end{table*}

\begin{figure*}[t]
    \centering
    \includegraphics[width=0.16\linewidth]{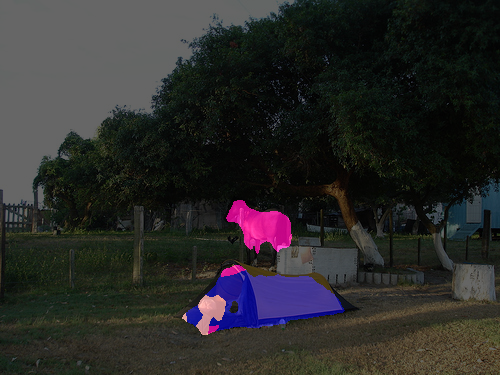}
    \includegraphics[width=0.16\linewidth]{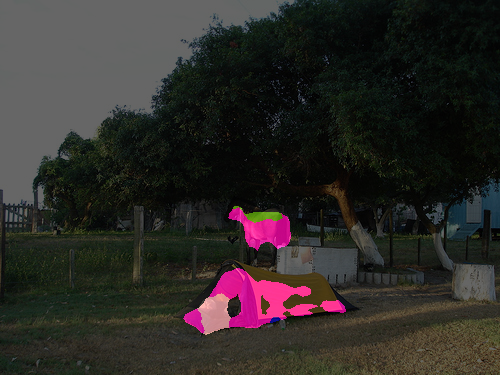}
    \includegraphics[width=0.16\linewidth]{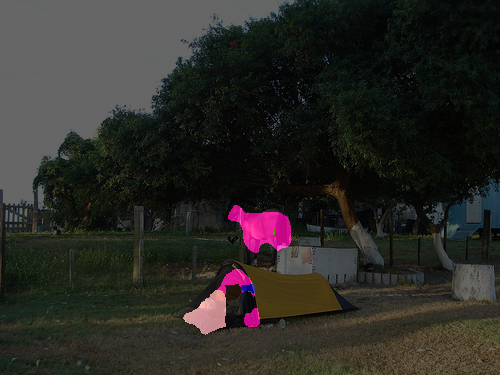}
    \includegraphics[width=0.16\linewidth]{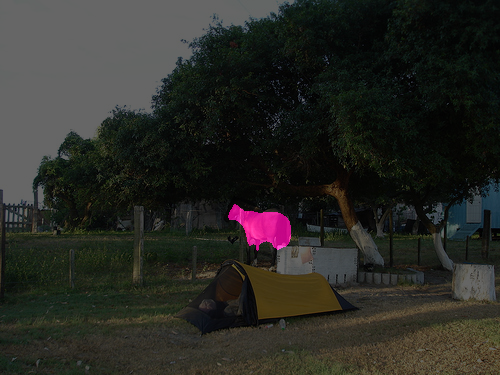}
    \includegraphics[width=0.16\linewidth]{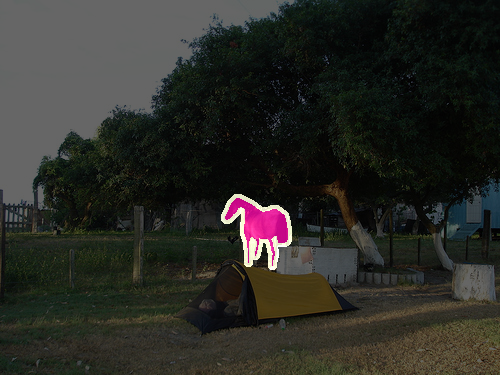}

    \includegraphics[width=0.16\linewidth]{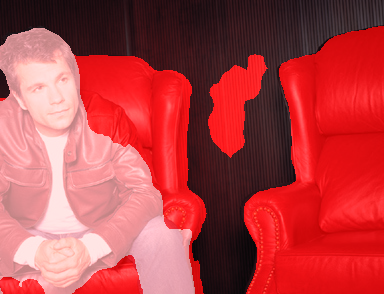}
    \includegraphics[width=0.16\linewidth]{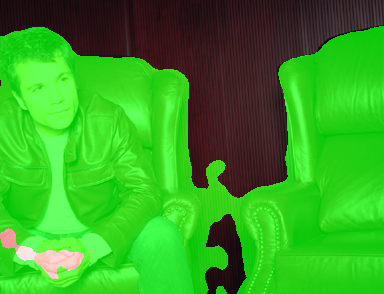}
    \includegraphics[width=0.16\linewidth]{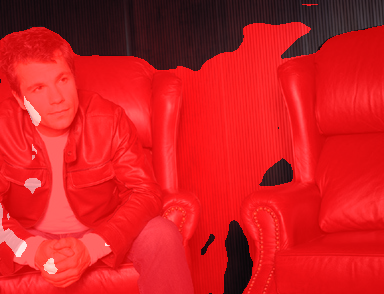}
    \includegraphics[width=0.16\linewidth]{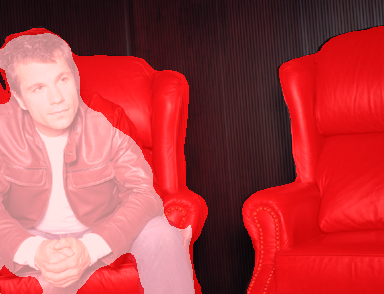}
    \includegraphics[width=0.16\linewidth]{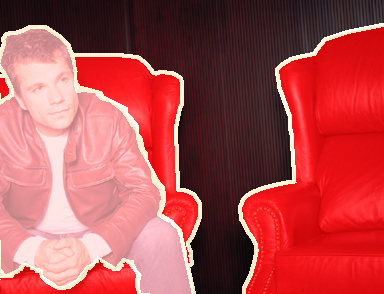}

    \includegraphics[width=0.16\linewidth]{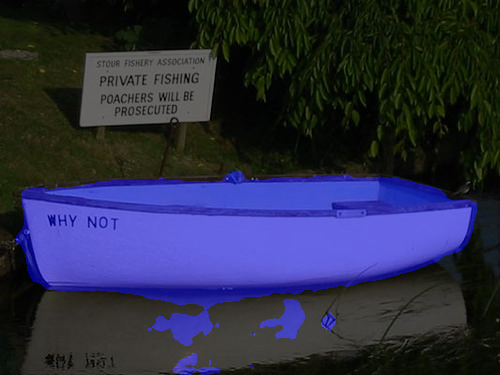}
    \includegraphics[width=0.16\linewidth]{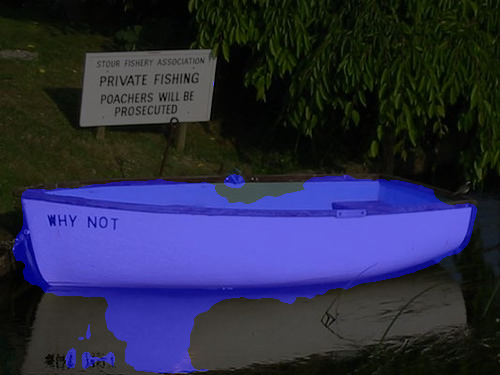}
    \includegraphics[width=0.16\linewidth]{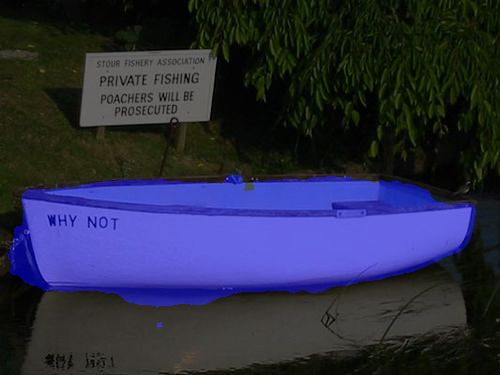}
    \includegraphics[width=0.16\linewidth]{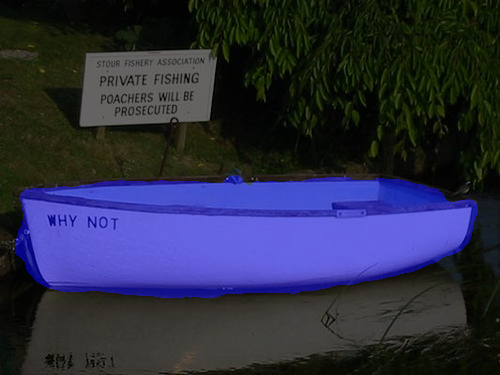}
    \includegraphics[width=0.16\linewidth]{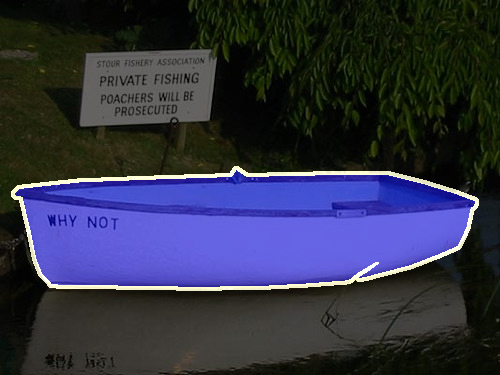}

    \includegraphics[width=0.16\linewidth]{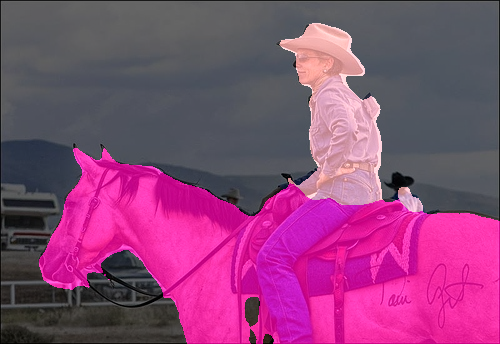}
    \includegraphics[width=0.16\linewidth]{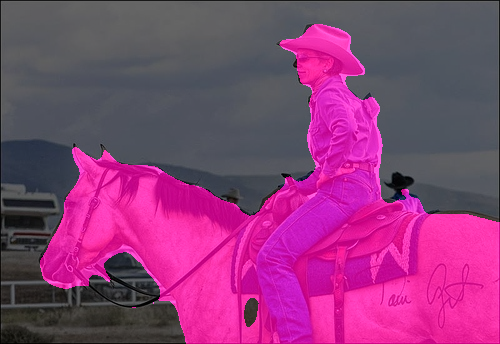}
    \includegraphics[width=0.16\linewidth]{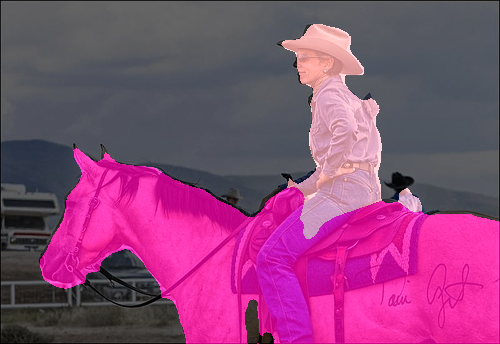}
    \includegraphics[width=0.16\linewidth]{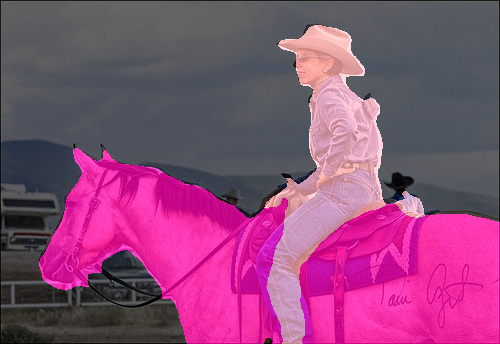}
    \includegraphics[width=0.16\linewidth]{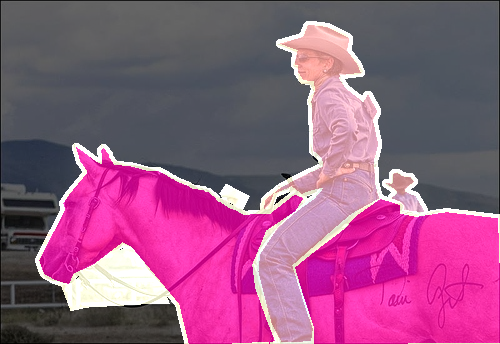}

    \includegraphics[width=0.16\linewidth]{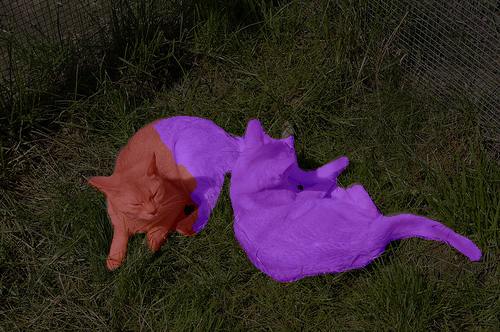}
    \includegraphics[width=0.16\linewidth]{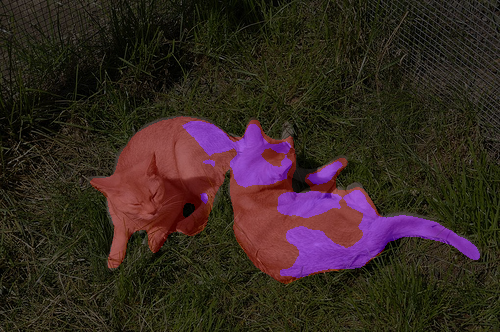}
    \includegraphics[width=0.16\linewidth]{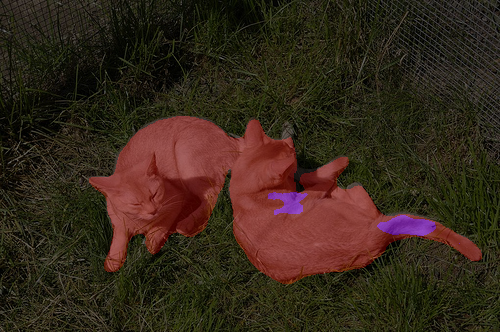}
    \includegraphics[width=0.16\linewidth]{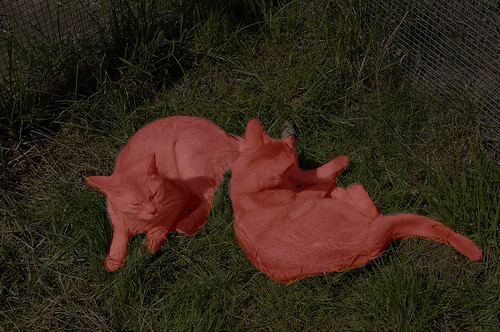}
    \includegraphics[width=0.16\linewidth]{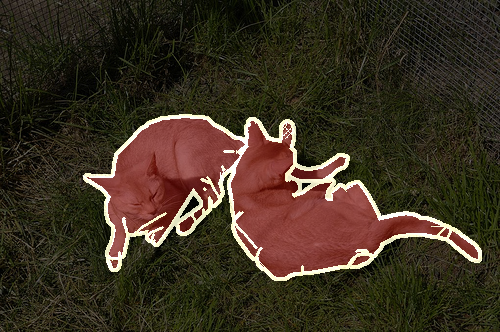}

      \quad\; w/o DLP \quad\quad\quad\quad\; w/o $\mathcal{L}^{Der}$   \quad\quad\; w/o momentum \quad\quad\quad\quad Ours \quad\quad\quad\quad\quad\;\;\; GT \quad\quad\;
    
    \caption{Visual results of ablation studies on Pascal VOC 2012.}
\label{com_ablation}
\end{figure*}

\begin{table*}[t]
\centering
\begin{tabular}{c|c|c|ccc}
\toprule
Method & Year & Input Size &  1/16 (186) & 1/8 (372) & 1/4 (744)  \\
\midrule
UniMatch & CVPR'23 &  $321^2$ &   58.6 & 58.7 &   60.8 \\
CorrMatch & CVPR'24 &  $321^2$ &    59.3 & 59.9  & 60.2 \\
ScaleMatch & AAAI'25 &  $321^2$ &  59.7 & 60.5 & 61.6  \\
\midrule
DerProp w/o DLP & - & $321^2$  &  59.1 & 60.4 & 60.9 \\
DerProp w/o $\mathcal{L}^{Der}$ & - & $321^2$  &  57.1 & 60.5 & 61.0 \\
DerProp w/o momentum & - & $321^2$  & 57.6 & 60.4 &  61.6 \\
\midrule
DerProp (Ours)  & - & $321^2$  & \textbf{60.1} & \textbf{61.8} &  \textbf{62.1} \\
\bottomrule
\end{tabular}
\caption{Results on Cityscapes in terms of mIoU (\%), with ResNet101 as the backbone. Best results are \textbf{Bolded}.}
\label{cityscapes_result}
\end{table*}

\subsection{Datasets \& Evaluation Metrics}
\label{datasets}
Two datasets are employed in this paper, which include: 1) the commonly used Pascal VOC 2012  \cite{PASCAL_VOC} having 1,464 training samples with 21 categories, and 1,449 testing samples, and 2) the urban scene understanding dataset Cityscapes \cite{cityscape} consisting of 2,975 training images and 500 validating images.

As for the evaluation metrics, we report mean Intersection-over-Union (mIoU) on the Pascal VOC 2012 validation set using original images, consistent with prior work \cite{DBLP:conf/cvpr/ChenYZ021, DBLP:conf/bmvc/FrenchLAMF20, MeanTeacher3, CorrMatch}. For Cityscapes, following \cite{DBLP:conf/cvpr/ChenYZ021,DBLP:conf/cvpr/WangWSFLJWZL22,UniMatch}, we evaluate via sliding window with fixed-size crops and compute mIoU on these crops. All results are obtained by using single-scale inference on the standard validation set.

\subsection{Ablation Study}
\label{sec:ablation}
To assess the influences of our proposed components, we compare our method with the following alternatives: 1) \textbf{w/o DLP}. In this setting, we train the segmentation network without using our proposed derivative label propagation. The predictions from the weakly augmented inputs are taken as the pseudo-labels for supervising the predictions from the corresponding strongly augmented images, without being rectified through Eq. \eqref{eq:pseudolabel_refine}. The derivative loss $\mathcal{L}^{Der}$ formulated in Eq. \eqref{loss:derivative} is also omitted; 2) \textbf{w/o $\mathcal{L}^{Der}$}. For this alternative, the pseudo-labels are rectified through Eq. \eqref{eq:pseudolabel_refine}, but the loss $\mathcal{L}^{Der}$ for regularizing the similarity matrix is abandoned; 3) \textbf{w/o momentum}. We do not maintain any momentum network. In testing, the back-propagation network $\mathcal{G}(\cdot)$ trained after the last epoch, is executed. In this setting, the pseudo-labels are rectified through  Eq. \eqref{eq:pseudolabel_refine}, and the loss $\mathcal{L}^{Der}$ is introduced. It is worth mentioning that, for both \textbf{w/o DLP} and \textbf{w/o $\mathcal{L}^{Der}$}, the momentum network is not used. For fair comparison, the performance of all the alternatives are evaluated using the same random seed, and the checkpoints saved at the same epoch. The network architectures and all other hyper-parameters are also kept to be the same.  Note that under the fully supervised scenario with all samples being labeled, \textit{i.e.}, \textbf{Full (1464)}, no pseudo-labels are required in training. Consequently, the \textbf{w/o DLP} variant is not applicable and its results are not reported for this data setting.

As given in Tab. \ref{PASCALVOC_result}  and Tab. \ref{cityscapes_result},  omitting our proposed derivative label propagation  induces significant performance degradation across all labeled data proportions, \textit{i.e.}, 1/16(92), 1/8(183), 1/4(366), and 1/2(732).  This underscores DLP’s efficacy in enhancing segmentation accuracy by rectifying  pseudo-labels.   Moreover, abandoning the derivative loss $\mathcal{L}^{Der}$ consistently reduces accuracy, confirming the necessity of this loss function for effective regularization. We also evaluate the performance of eliminating  the momentum network in testing, which  yields measurable performance deterioration. It highlights the role in maintaining satisfactory segmentation accuracy. 

Fig. 3 provides visual results to reveal that  the proposed derivative label propagation is effective for improving the segmentation performance. It can be observed that, our method exhibits precise boundary delineation and structural integrity, particularly for small objects.

\subsection{Comparisons with other State-of-the-arts}
\label{sec:com}
We compare our DerProp with recent semi-supervised segmentation works, which include: UniMatch \cite{UniMatch}, AllSpark \cite{AllSpark}, RankMatch \cite{RankMatch}, CorrMatch \cite{CorrMatch}, MGCT \cite{MGCT}, RCC \cite{RCC}, ScaleMatch \cite{ScaleMatch}. Note that previous semi-supervised semantic segmentation methods use an input size of $801\times 801$ on the Cityscapes dataset, while our method uses a smaller size of $321\times 321$. Hence, for fair comparison, we re-train other competitors, \textit{i.e.}, UniMatch, CorrMatch, and ScaleMatch, with $321\times 321$ inputs. 

\begin{figure}[t]
    \centering
    \includegraphics[width=0.27\linewidth]{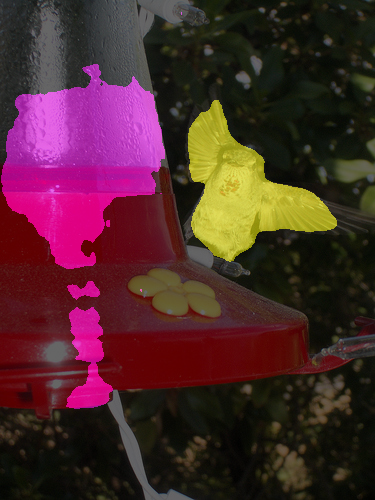}
    \includegraphics[width=0.27\linewidth]{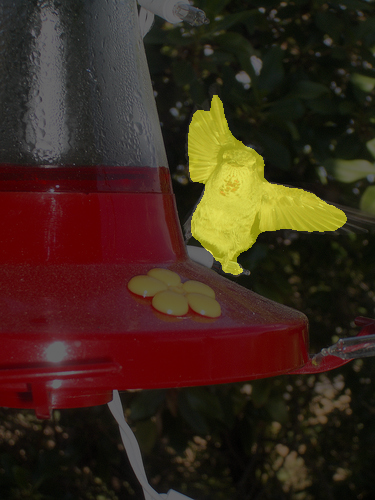}
    \includegraphics[width=0.27\linewidth]{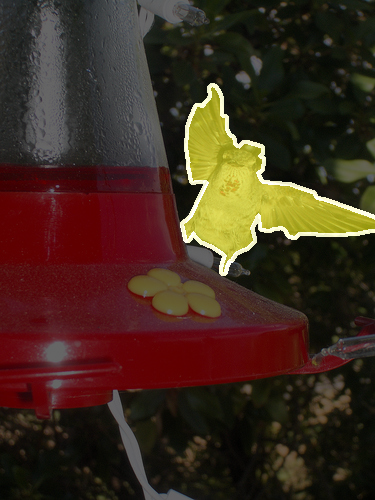}

    \includegraphics[width=0.27\linewidth]{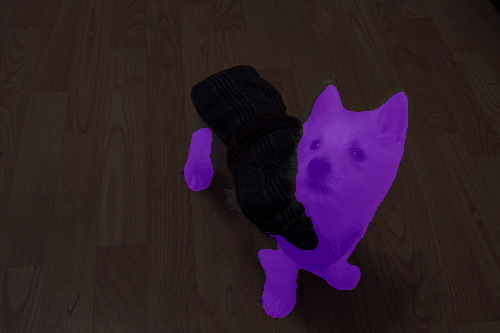}
    \includegraphics[width=0.27\linewidth]{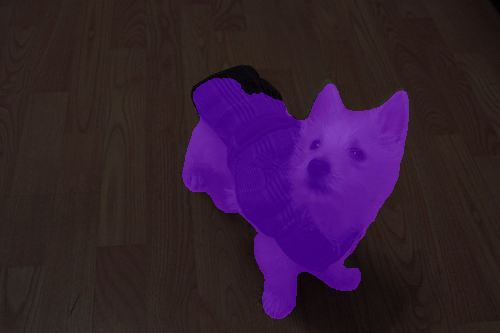}
    \includegraphics[width=0.27\linewidth]{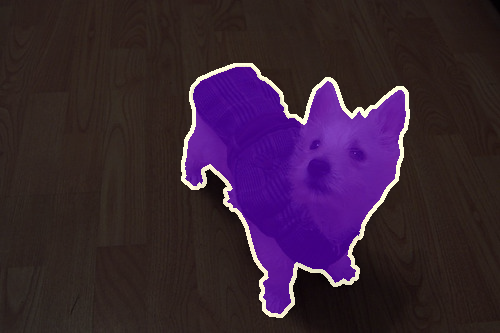}

    \includegraphics[width=0.27\linewidth]{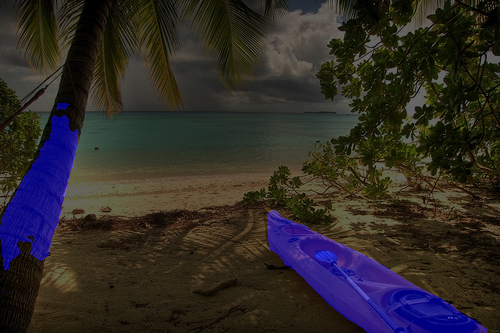}
    \includegraphics[width=0.27\linewidth]{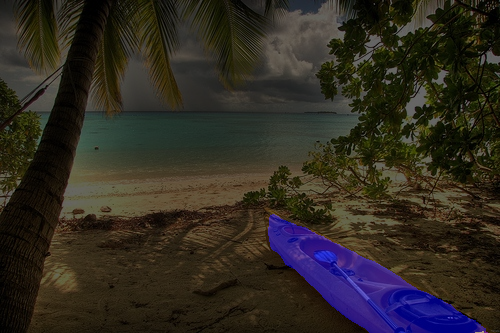}
    \includegraphics[width=0.27\linewidth]{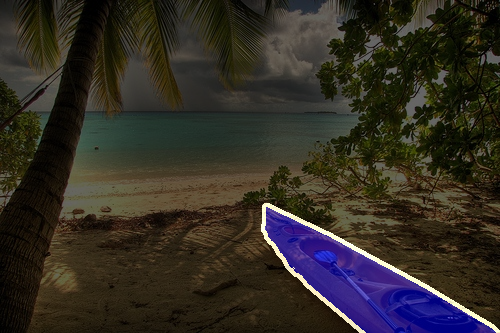}

    \includegraphics[width=0.27\linewidth]{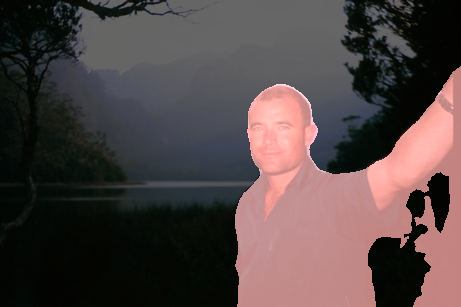}
    \includegraphics[width=0.27\linewidth]{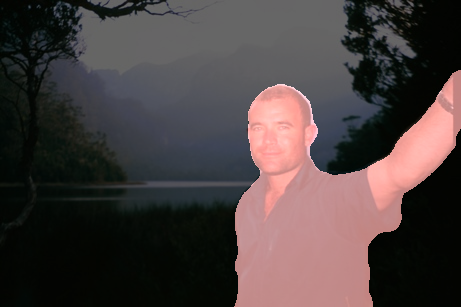}
    \includegraphics[width=0.27\linewidth]{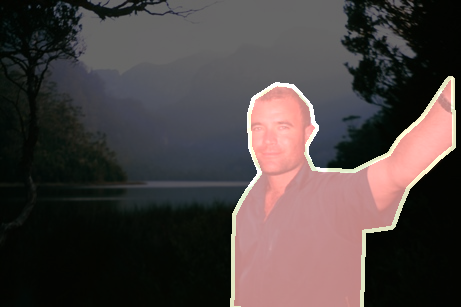}

      CorrMatch \quad\quad\quad\; Ours \;\quad\quad\quad\quad GT \quad\quad
    
    \caption{Visual comparisons on Pascal VOC 2012.}
\label{com}
\end{figure}

As given in Tab. \ref{PASCALVOC_result}, our method achieves state-of-the-art performance across all labeled data proportions, \textit{i.e.}, 1/16(92), 1/8(183), 1/4(366), and 1/2(732), surpassing recent competitors by significant margins.  This confirms that our proposed DLP remains effective with different number of labeled and unlabeled samples. For example, DerProp outperforms CorrMatch (76.4\%) and ScaleMatch (76.1\%) by around 1\%, which demonstrates our exceptional robustness to extreme label scarcity. While methods like  AllSpark use larger input size, \textit{i.e.}, $513\times 513$, and larger amount of parameters, our DerProp still achieves superior accuracy with smaller $321\times 321$ inputs  without introducing additional  parameters, highlighting its computational efficiency.
As for the Cityscapes dataset, it can be inferred from Tab. \ref{cityscapes_result} that, our DerProp consistently dominates all settings, achieving 62.1\% mIoU at 1/4 labeled data (744 images), surpassing ScaleMatch (61.6\%) and UniMatch (60.8\%). Notably, our method outperforms CorrMatch, which also uses correlation maps for label propagation, by a large margin, demonstrating that the derivative operations act as effective constraints for restricting the solution space.
Moreover, as shown in Fig. \ref{com},   DerProp eliminates undesired misclassifications by CorrMatch, \textit{e.g.}, in the last row, ``vegetation" is  mislabeled as ``person" by CorrMatch.

\subsection{Discussion on the Derivative Losses}
In this section, we evaluate the results of altering the supervisions on similarity matrices with respect to derivative features. Similar to the ablation studies, all the alternatives use the same random seed, and the checkpoints saved at the same epoch. Specifically, for $Q \in \{0, 1, ..., D-1\}$, we give the general form of derivative loss as:
\begin{equation}
\begin{aligned}
    \mc{L}_{Q}^{Der} &\defeq  \sum^{N_l}_{nl\defeq 1} [(\sum_{q\defeq 0}^Q ||\Delta^q \mathbf{S}_{nl} - \mathbf{S}_{nl}^{GT}||_1) +  \eta ||\Delta^{Q+1} \mathbf{V}_{nl}||_1] \\
   + & \sum_{nu\defeq 1}^{N_u} [(\sum_{q\defeq 0}^Q ||\Delta^q \mathbf{S}^s_{nu} - \Delta^q \mathbf{S}_{nu}^{GT}||_1) 
    +  \eta ||\Delta^{Q+1} \mathbf{V}^s_{nu}||_1],
    \end{aligned}
    \label{loss:derivative_reformulation}
\end{equation}
where   $\Delta^q\mathbf{S}_{nu}^{GT} \defeq (\Delta^q\mathbf{V}_{nu}^w)^{\top}(\Delta^q\mathbf{V}^w_{nu}) \in\mathbb{R}^{M\times M}$, $\forall q\in\{1, 2, ..., D-1\}$, and $\Delta^q \mathbf{V}^w_{nu}$ is the $q$-order derivative features projected by $\mathcal{J}(\cdot)$ corresponding to the $nu$-th weakly augmented image $\mathbf{X}^w_{nu}$. $\Delta^q\mathbf{S}_{nu}^{s}$ is calculated by $ (\Delta^q\mathbf{V}_{nu}^s)^{\top}(\Delta^q\mathbf{V}^s_{nu})$, and $\Delta^q \mathbf{V}^s_{nu}$ is the $q$-order derivative features corresponding to the $nu$-th strongly augmented image $\mathbf{X}^s_{nu}$.
When $Q\defeq 1$, the general form of derivative loss in Eq.   \eqref{loss:derivative_reformulation}, say $\mathcal{L}_{1}^{Der}$, is consistent with  the loss in Eq. \eqref{loss:derivative}.

As given in Tab. \ref{tab:DerivativeLoss}, the impact of varying the parameter $Q$ in Eq. \eqref{loss:derivative_reformulation} on segmentation accuracy is evident. A clear trend emerges: increasing $Q$ results in a significant degradation in performance. Consequently, $\mathcal{L}_{1}^{Der}$ ($Q=1$) demonstrates superior effectiveness compared to other alternatives with higher $Q$ values. Furthermore, we observe a consistent accuracy drop across all evaluated derivative losses, \textit{i.e.}, $\mathcal{L}_{0}^{Der}$, $\mathcal{L}_{1}^{Der}$, $\mathcal{L}_{2}^{Der}$, and $\mathcal{L}_{3}^{Der}$, when the sparsity regularization terms, \textit{i.e.}, $\eta||\Delta^{Q+1}\mathbf{V}_{nl}||_{1}$ and $\eta||\Delta^{Q+1}\mathbf{V}_{nu}^{s}||_{1}$ in Eq. \eqref{loss:derivative_reformulation}, are omitted (see the results of \textbf{w/o sparsity}  in Tab. \ref{tab:DerivativeLoss}). This pronounced decline validates the efficacy of the proposed sparsity regularization in constraining the solution space and enhancing segmentation accuracy.

\subsection{Discussion on the Derivative Operations}
In this section, we evaluate the results of replacing our proposed derivative operation formulated in Eq. \eqref{eq:derivative1} with the following three different operations:
\begin{equation}
\Delta^q \mathbf{v}(i) \defeq  (\Delta^{q-1} \mathbf{v}(i+2) - \Delta^{q-1} \mathbf{v}(i))/2,
\label{eq:derivative2}
\end{equation}
\begin{equation}
\Delta^q \mathbf{v}(i) \defeq  \Delta^{q-1} \mathbf{v}(i+1) + \Delta^{q-1} \mathbf{v}(i),
\label{eq:derivative3}
\end{equation}
\begin{equation}
\Delta^q \mathbf{v}(i) \defeq  \Delta^{q-1} \mathbf{v}(i+1) + \Delta^{q-1} \mathbf{v}(i-1) - 2\Delta^{q-1} \mathbf{v}(i).
\label{eq:derivative4}
\end{equation}
As given in Tab. \ref{tab:DerivativeOperation}, our proposed derivative operation in Eq. 2 consistently outperforms all the other alternatives. To be more specific, Eq. \eqref{eq:derivative2} achieves suboptimal results compared with our proposed derivative operation, due to its sensitivity to feature noises. Compared with our method, the wider receptive field (from the $i$-th to $(i + 2)$-th element of features vectors) of Eq. \eqref{eq:derivative2}  introduces instability when capturing high-frequency semantic transitions. The summation operation formulated in Eq. \eqref{eq:derivative3} performs the worst among all the alternatives, as it fails to capture feature variations. It solely accumulates adjacent feature values, contradicting the derivative’s purpose of modeling semantic differences between adjacent elements of feature vectors. Moreover, Eq. \eqref{eq:derivative4} shows marginal improvement over Eq. \eqref{eq:derivative3}, but remains inferior to Eq. \eqref{eq:derivative2}. Its operation over-constrains the solution space, suppressing subtle class boundaries. To ensure the
fairness, we also use the checkpoints saved at the same epoch for all the alternatives.

Generally, our derivative operation in Eq.  \eqref{eq:derivative2}, optimally balances noise robustness and discriminative capacity.  According to \textbf{Theorem} \ref{theorem1}, our subtraction calculation between adjacent elements,  effectively preserves the semantics of pixels while mitigating the ill-posedness.

\begin{table}[t]
\centering
\begin{tabular}{c|cc}
\toprule
Method & w/o sparsity & w/ sparsity  \\
\midrule
$\mathcal{L}^{Der}_{0}$ & 74.1 &  75.1 \\
$\mathcal{L}^{Der}_{1}$ (Ours) & \textbf{77.0} & \textbf{77.6} \\
$\mathcal{L}^{Der}_{2}$ & 76.0 & 76.9 \\
$\mathcal{L}^{Der}_{3}$ & 76.4 & 76.7 \\
\bottomrule
\end{tabular}
\caption{Results of different derivative losses on Pascal VOC 2012 in terms of mIoU (\%). All the results in this table are obtained under the setting of 1/16 (92), \textit{i.e.}, the amount of labeled images is 92, which accounts for 1/16 of total training samples. Best results are \textbf{Bolded}.}
\label{tab:DerivativeLoss}
\end{table}

\begin{table}[t]
\centering
\begin{tabular}{c|cc}
\toprule
Method & 1/16 (92) & 1/8 (183)  \\
\midrule
Eq. \eqref{eq:derivative2} & 76.1 &  78.6 \\
Eq. \eqref{eq:derivative3} & 76.4 & 77.1 \\
Eq. \eqref{eq:derivative4} & 76.8 & 78.5 \\
Eq. \eqref{eq:derivative1} (Ours) & \textbf{77.6} & \textbf{78.7} \\
\bottomrule
\end{tabular}
\caption{Results of different derivative operations on Pascal VOC 2012 in terms of mIoU (\%). Best results are \textbf{Bolded}.}
\label{tab:DerivativeOperation}
\end{table}

\section{Conclusion}
We proposed DerProp, a novel semi-supervised semantic segmentation framework that addresses the  challenge of unreliable pseudo-labels through derivative label propagation (DLP). Our  innovation lies in imposing discrete derivative operations on pixel-wise feature vectors, which introduces additional regularization on similarity metrics. As a profit, the proposed DLP effectively alleviates the ill-posed problem where identical similarities correspond to divergent features. 
Extensive experiments were conducted on Pascal VOC 2012 and Cityscapes dataset to demonstrate the state-of-the-art performance of our proposed SSSS method.

\section{Theoretical Evidences}

\subsection{Proof of Theorem 1}
\label{sec:proof1}
The $q$-order derivative operator can be represented as a matrix multiplication:
\begin{equation}
    \Delta^q \mathbf{v} = \mathbf{A}_q \mathbf{v},
\end{equation}
where $\mathbf{v}\in \mathbb{R}^{D}$ is a $D$-dimensional column vector, and $\mathbf{A}_q \in \mathbb{R}^{(D-q) \times D}$ is the Toeplitz matrix:
\begin{equation}
    \mathbf{A}_Q(i,j) = 
    \begin{cases}
        (-1)^{q-p} \binom{q}{p} & \text{if } j = i + p, \, 0 \leq p \leq q \\
        0 & \text{otherwise}
    \end{cases}.
    \label{eq:derivativeOperation}
\end{equation}
Then, $\forall q \in \{1, 2, ..., D - 1\}$, this regularization forms non-linear equations:
\begin{align}
    \mathcal{S}(\mathbf{v}_{i}, \mathbf{v}_{j}) \defeq \frac{\mathbf{v}^\top_{i} \mathbf{v}_{j}}{\|\mathbf{v}_{i}\| \|\mathbf{v}_{j}\|} &= s_{i,j}, \label{eq:original} \\
    \mathcal{S}(\Delta^q \mathbf{v}_{i}, \Delta^q \mathbf{v}_{j}) \defeq \frac{(\mathbf{A}_q \mathbf{v}_{i})^\top (\mathbf{A}_q \mathbf{v}_{j})}{\|\mathbf{A}_q \mathbf{v}_{i}\| \|\mathbf{A}_q \mathbf{v}_{j}\|} &= s_{i,j}^{q}, \label{eq:derivative}
\end{align}
where $\top$ is the matrix transpose operation. Under normalization condition $\|\mathbf{v}_{i}\| = \|\mathbf{v}_{j}\| = 1$, the non-linear terms $\|\mathbf{v}_{i}\|$ and $\|\mathbf{v}_{j}\|$ can be omitted. Further, given that $\mathbf{A}_q$ is a linear transformation, Eqs. ~\eqref{eq:original} and ~\eqref{eq:derivative} can be simplified to the system of linear equations:
\begin{equation}
    (\mathbf{A}\mathbf{v}_{i})^\top(\mathbf{A}\mathbf{v}_{j})=\mathbf{S},
\end{equation}
where $\mathbf{A}$ is a joint coefficient matrix, and $\mathbf{S}$ is the column vector of cosine similarities. $\mathbf{A}$ and $\mathbf{S}$  are formulated as:
\begin{equation}
    \mathbf{A} \defeq \left[
\begin{matrix}
\mathbf{I} \\
\mathbf{A}_1 \\
\vdots \\
\mathbf{A}_{D-1}
\end{matrix}  
\right] \in \mathbb{R}^{[D+(D-1)+...+1]\times D};  \mathbf{S} \defeq \left[
\begin{matrix}
s_{i,j} \\
s_{i,j}^{1} \\
\vdots \\
s_{i,j}^{D-1}
\end{matrix}  
\right] \in \mathbb{R}^{D},
\label{eq:jointcoefficient}
\end{equation}
where $\mathbf{I}\in\mathbb{R}^{D\times D}$ is an identity transformation matrix, corresponding  to the original features $\mathbf{v}$ without any derivative operation.  The rank of each sub-matrix in $\mathbf{A}$ is:
\begin{equation}
    \text{Rank}(\mathbf{I}) = D, \text{Rank}(\mathbf{A}_q) = D - q, \forall q \in \{1, 2, ..., D-1\}.
\end{equation}
The rows in each matrix $\mathbf{A}_q, q \in\{1, 2, ..., D-1\}$, for derivative operation, are essentially a linear combination of the rows in $\mathbf{I}$, which is:
\begin{equation}
\mathbf{A}_q(i) \defeq \sum_{d\defeq 1}^D c^q_{i,d} \; \mathbf{I}(d),
\end{equation}
where $\mathbf{A}_q(i)$ and $\mathbf{I}(d)$ mean the $i$-th and $d$-th row of $\mathbf{A}$ and $\mathbf{I}$, respectively. $c^q_{i,d}$ is a scalar determined by the order of derivative operations. Because the rank of a joint matrix is calculated by selecting the maximum number of linearly-invariant rows, we have:
\begin{equation}
\text{Rank}(\mathbf{A}) \defeq \text{Rank}(\mathbf{I}) = D.
\end{equation}

Since the rank of $\mathbf{A}$ is equal to its number of columns, \textit{i.e.}, $D$, $\mathbf{A}$ is full rank. The  uniqueness in the system of linear equations reveals that, if the joint coefficient matrix $\mathbf{A}$ is full rank, there exists a unique solution in solving the linear equations. Hence, there exists a unique solution of $D$-dimensional feature vectors, given $D$ constraints.

Proof completed.

\subsection{Proof of Theorem 2}
To begin with,  according to \textbf{Lemma} \ref{lemma:L1Norm}, $\forall q \in\{2, ..., D-1\}$, we have:
\begin{equation}
\begin{aligned}
   & ||\Delta^q\mathbf{v}||_1 \leq 2^{q-2}\cdot ||\Delta^2\mathbf{v}||_1 \\
    \Rightarrow & ||\Delta^q\mathbf{V}||_1 \leq 2^{q-2}\cdot ||\Delta^2\mathbf{V}||_1.
    \end{aligned}
\end{equation}
Hence, for the norm of similarity matrix, we can obtain:
\begin{equation}
\begin{aligned}
   ||\Delta^q\mathbf{S}||_1 \leq & (\sum^M_{i\defeq 1} ||\Delta^q\mathbf{v}_i||_1)^2 \\
   = & (||\Delta^q\mathbf{V}||_1)^2  \leq (2^{q-2}\cdot ||\Delta^2\mathbf{V}||_1)^2.
    \end{aligned}
\end{equation}
Proof completed.

\subsection{Lemma for L1 Norms of Derivative Operations}
\begin{lemma}
Given that $\Delta^q \mathbf{v} \defeq\mathbf{A}_{q} \mathbf{v}$, the following inequality holds:
\begin{equation}
\begin{aligned}
         ||\Delta^q \mathbf{v}||_1 \leq ||\mathbf{A}_q||_{1\rightarrow 1} \cdot||\mathbf{v}||_1 = 2^q \cdot ||\mathbf{v}||_1,\\
\end{aligned}
\end{equation}
where $||\mathbf{A}_q||_{1\rightarrow1} \defeq \max_{1 \leq j \leq D} \sum_{i\defeq 1}^{D-q} |\mathbf{A}_q(i,j)|$, and $\mathbf{v} \in \mathbb{R}^D$ is a $D$-dimensional vector.
As formulated in Eq. \eqref{eq:derivativeOperation}, the non-zero elements in each row of $\mathbf{A}_q$ are binomial coefficients. Since the sum of the absolute values of the binomial coefficients is $2^q$, it can be concluded that: $||\mathbf{A}_q||_{1\rightarrow1} \defeq 2^q$.
\label{lemma:L1Norm}
\end{lemma}

\bibliography{aaai2026}

\end{document}